\pdfoutput=1

\documentclass[11pt]{article}

\usepackage{ACL2023}

\usepackage{times}
\usepackage{latexsym}

\usepackage[T1]{fontenc}

\usepackage[utf8]{inputenc}

\usepackage{microtype}

\usepackage{inconsolata}

\usepackage{amsmath, amssymb}
\usepackage{multirow}
\usepackage{booktabs}
\usepackage{tikz-dependency}
\usepackage{pgfplots}
\usepackage{dashrule}
\usepackage{booktabs,arydshln}
\usepackage{scalerel}
\usepackage{booktabs, tabularx}
\usepackage{graphics}
\usepackage{multirow}
\usepackage{amsmath}
\usepackage{array,multirow,graphicx}
\definecolor{skyblue}{RGB}{70, 130, 180}
\definecolor{redhi}{RGB}{248,206,204}

\newcommand*{\affmark}[1][*]{\textsuperscript{#1}}
\makeatletter
\def\thanks#1{\protected@xdef\@thanks{\@thanks
        \protect\footnotetext{#1}}}
\makeatother

\title{Better Sampling of Negatives \\for Distantly Supervised Named Entity Recognition}
\author{Lu Xu\affmark[* 1 2 ]\thanks{$^*$ This work was done when Lu Xu was under the joint Ph.D. program between Alibaba and SUTD.} ~~~Lidong Bing\affmark[1] ~~~Wei Lu\affmark[2] \\$^1$DAMO Academy, Alibaba Group~~\\
$^2$Singapore University of Technology and Design \\
\texttt{xu\_lu@hotmail.com} ~~~ \texttt{l.bing@alibaba-inc.com}\\
\texttt{luwei@sutd.edu.sg}\\
}

\begin{document}
\maketitle
\begin{abstract}
Distantly supervised named entity recognition (DS-NER) has been proposed to exploit the automatically labeled training data instead of human annotations. 
The distantly annotated datasets are often noisy and contain a considerable number of false negatives. 
The recent approach uses a weighted sampling approach to select a subset of negative samples for training. 
However, it requires a good classifier to assign weights to the negative samples. In this paper, we propose a simple and straightforward approach for selecting the top negative samples that have high similarities with all the positive samples for training. 
Our method achieves consistent performance improvements on four distantly supervised NER datasets. Our analysis also shows that it is critical to differentiate the true negatives from the false negatives.\footnote{Our code is available at \url{https://github.com/xuuuluuu/ds_ner}.}
\end{abstract}

\section{Introduction}
\label{sec:intro}
Named entity recognition (NER) is one of the fundamental tasks in natural language processing, and it aims to extract the mentioned entities in the text.
Existing supervised approaches \cite{lample2016neural, Ma2016EndtoendSL, devlin2019bert, Xu2021BetterFI} have achieved great performance on many NER datasets. However, they still heavily rely on the human-annotated training datasets.

Distantly supervised approaches \cite{10.1145/2783258.2783362, Fries2017SwellSharkAG, Shang2018LearningNE,yang-etal-2018-distantly,DBLP:conf/conll/MayhewCTR19,Cao2019LowResourceNT,peng-etal-2019-distantly,Liu2021NoisyLabeledNW,zhang-etal-2021-de,zhou-etal-2022-distantly} have been proposed to exploit the automatically labeled training data generated from the knowledge bases (KBs) or dictionaries. 
For such distantly supervised datasets, the annotated entities mostly have correct labels, but the overall annotations are frequently incomplete due to the limited coverage of entities in KBs.
We include comparisons of the distantly-annotated and human-annotated datasets in Appendix \ref{app:stats}.

\begin{figure}[t]
\vspace{-.25cm}
    \centering
    \includegraphics[scale=0.4]{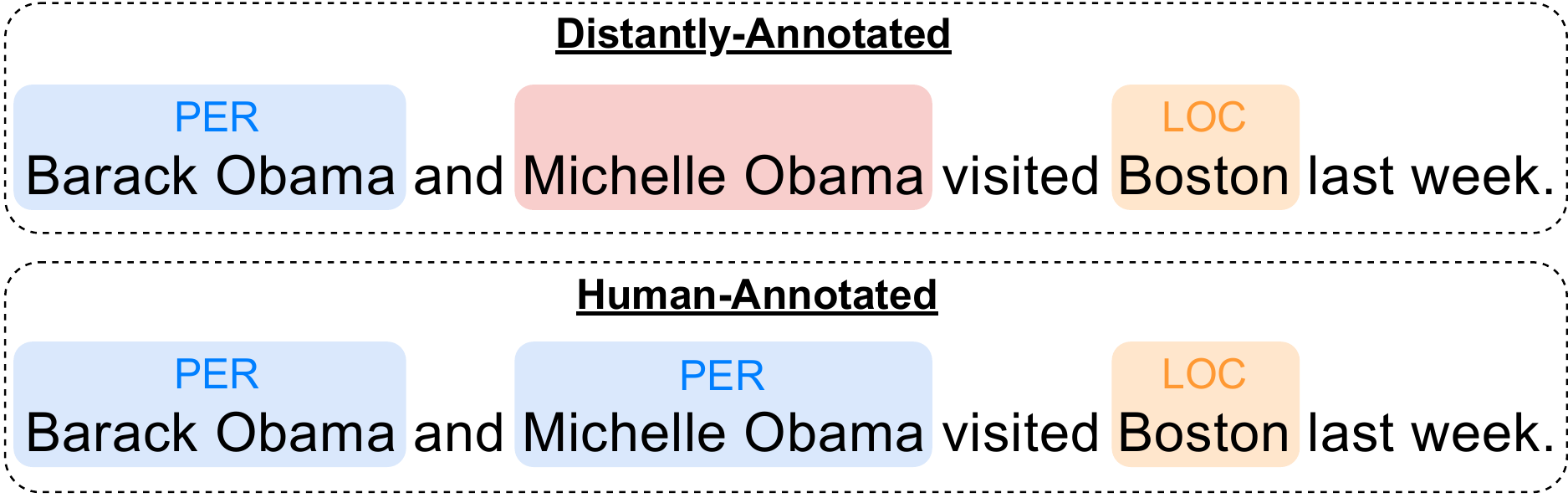}
    \vspace{-.65cm}
    \caption{An annotated example with distant supervision. The  entity highlighted in \colorbox{redhi}{red} is not recognized.}
    \label{fig:example}
    \vspace{-.45cm}
\end{figure}

Self-training has been demonstrated as an effective strategy for addressing the noisy labeled training data \cite{jie-etal-2019-better, bond, zhang:2021, DBLP:conf/emnlp/0001ZHWZJ021, tan-etal-2022-revisiting}. 
Specifically, they iteratively refine the entity labels through teacher-student models.
In this way, the number of false positive and false negative samples can be reduced.
However, such approaches usually require training multiple models with multiple iterations.
Another line of work improves the single-stage model by reducing the number of false negative samples during training. In general, the problem of false negatives is more severe than false positives.
\citet{li2021empirical} proposed to sample a portion of negative samples with a uniform sampling distribution for training. 
By selecting a subset of all the negative samples,  fewer false negative samples are involved in the training process.
The sampling strategy can be enhanced by using a weighted sampling distribution~\cite{li-etal-2022-rethinking}. Specifically, the negative samples are assigned with different sampling probabilities based on the predicted label distributions.
However, this approach also depends on quality of the classifier to 
derive the distribution.

In this paper, we propose a simple and straightforward approach to sampling the negatives for training.
Intuitively, the false negatives are positive samples but are unrecognized based on distant supervision, and they should have high similarities with positive samples that have the same gold entity type. 
For the example in Fig. \ref{fig:example}, 
``Michelle Obama'' is not identified as \textit{PER} in the distantly labeled dataset. 
The false negative sample ``Michelle Obama'' should have high similarity with the positive sample ``Barack Obama'' that has the \textit{PER} label.
Additionally, the false negative sample should not have high similarity with other positive samples of different entity types, such as ``Boston''. 
Therefore, when the negative samples have high similarities with all the positive samples, they are more likely to be true negatives.
We select these top negative samples for training, and we denote our approach as \textbf{Top-Neg}. 
Unlike the previous approach of relying on a classifier for assigning sampling probability, our approach exploits the encoded representations to derive the similarity scores.
Compared to the baseline methods, our approach demonstrates consistent performance improvement on four distantly supervised NER datasets. Our analysis shows that not all the negative samples are required for training, but it is critical to filter the false negatives.

\section{Approach}

The objective of NER task is to extract all the entities in the text. 
Given a sentence of length $n$, $X = \{x_1, x_2, ..., x_n\}$, 
we denote all possible enumerated spans in $X$ as $S = \{s_{1,1}, \; s_{1,2}, \; ..., \; s_{i,j}, \; ..., \; s_{n,n}\}$, where $i$ and $j$ are the start and end position of span $s_{i,j}$, and the span length is limited as $0 \leq j-i \leq L$.
For all the enumerated spans, our approach predicts the corresponding entity types from a predefined label space, including the \textit{O} label (not an entity).

\subsection{Span-based Model}
Similar to the previous  approaches \cite{lee-etal-2017-end, luan-etal-2019-general, zhong2021frustratingly, xu-etal-2021-learning}, we adopt a span-based model architecture.
First, we encode the input sentence $X$ with a pre-trained language model, such as BERT \cite{devlin2019bert}. The encoded contextualized representation for the sentence $X$ is denoted as $\mathbf{h}= [ \mathbf{h}_1, \mathbf{h}_2, ..., \mathbf{h}_n]$. 
Then, the span representation of $s_{i,j} \in S$ can be formed as:
\begin{equation}
\label{spanrep}
    \mathbf{s}_{i,j} =   
        [\mathbf{h}_i ; \; \mathbf{h}_j ;\;
        f(i,j)]
\end{equation}
where $f(i,j)$ indicates a trainable embedding to encode the span width feature.
The span representation $\mathbf{s}_{i, j}$ is then input to a feed-forward neural network (FFNN) to obtain the distribution of entity type $t$. 
\begin{equation}
\label{eq:classifier}
P(t|\mathbf{s}_{i, j}) = \mathrm{softmax}(\mathrm{FFNN} (\mathbf{s}_{i, j}))
\end{equation}

\subsection{True Negatives vs. False Negatives}
\label{sec:ours}
In general, the distantly annotated training datasets contain a significant number of false negative samples and also a small portion of false positives. 
When the model is trained on such datasets, the performance of precision and recall are affected, and the preliminary experimental results are given in Appendix \ref{app:preliminary}. We observe that the recall score is severely affected when compared with the precision score. Such behavior is because the problem of false negatives is more severe than false positives. We also observe a similar phenomenon in the statistics of the datasets in Appendix \ref{app:stats}. 

Regarding the false negative samples, they are actually true positives but cannot be annotated based on only the distantly supervised information.
Through our intuition that is described in Section~\ref{sec:intro}, the false negative samples should have high similarities with the positive samples having the same gold entity type and low similarities with other positive samples of different entity types. 
Note that a vanilla model that is trained on the distantly supervised dataset can still well differentiate the positive labels, as demonstrated by the high precision score in the Appendix \ref{app:preliminary}.
With the above findings, when a negative sample has a high similarity with all the positive samples, it is likely to be a true negative sample. 
Therefore, we propose to only utilize the negative samples that have high similarities with all the positive samples for training. 

At the training stage, we have the label information so that we can obtain the span set of positive samples $S^{pos} = \{..., s^{pos}, ...\}$, and span set of negative samples $S^{neg} = \{..., s^{neg}, ...\}$. Note that $S = S^{pos} \cup S^{neg}$.
Then, we calculate the average similarity score of each negative span $s^{neg} \in S^{neg}$ with respect to all the positive spans in $S^{pos}$, and the similarity score $\Phi$ is defined as:
\begin{equation}
\label{eq:sim}
    \Phi (\mathbf{s}^{neg}, S^{pos}) = \frac{1}{M} \sum_{\mathbf{s}^{pos} \in S^{pos}}  \frac{\mathbf{s}^{neg}}{\|\mathbf{s}^{neg} \|} \cdot \frac{\mathbf{s}^{pos}}{\|\mathbf{s}^{pos} \|} 
\end{equation}
where $M$ denotes the number of positive samples. In practice, we calculate the similarity score at the batch level.

\subsection{Training and Inference}
We rank the similarity score $\Phi$ of all the negative spans in $S^{neg}$. Note that  the number of the negative samples is denoted as $N$, which has a complexity of $O(n^2)$.
To only consider the negative samples that have high similarities with all the positives and also save the computational cost, we select the top $Nr$ negative samples for training and $r$ is the hyper-parameter to control the quantity. 
We denote the set of the selected negative samples as $\tilde{S}^{neg} = \{..., \tilde{s}^{neg}, ...\}$.
Then, we input all the positive samples and the selected negative samples to Eq. \ref{eq:classifier} to obtain the probability distributions.
Our training objective is defined as:
\begin{equation}
\begin{split}
    \mathcal{L} =
    & - \sum_{\mathbf{s}^{pos} \in S^{pos}}\log P(t^* | \mathbf{s}^{pos}) \\
    & - \sum_{\tilde{\mathbf{s}}^{neg} \in \tilde{S}^{neg}}\log P(t^* | \tilde{\mathbf{s}}^{neg}) \\
\end{split}
\end{equation}
where $t^*$ denotes the corresponding gold entity type of a span.
During inference, all the enumerated span representations are passed to Eq. \ref{eq:classifier} to predict the corresponding entity types.

\section{Experiments}

\paragraph{Datasets} We evaluate our approach on four distantly supervised NER datasets: 
CoNLL03 \cite{tjong-kim-sang-de-meulder-2003-introduction}, BC5CDR \cite{bc5cdr}, WNUT16 \cite{godin-etal-2015-multimedia}, and WikiGold \cite{balasuriya-etal-2009-named}.
The distantly supervised datasets are obtained from \cite{bond} and \cite{Shang2018LearningNE}. We use the distantly supervised data for training and the human-annotated development and test sets for evaluation. The statistics of the datasets are given in Appendix~\ref{app:stats}.

\paragraph{Experimental Setup}
We use the \emph{bert-base-cased} and \emph{roberta-base} as the base encoders for CoNLL03, WNUT16, and WikiGold datasets. BC5CDR is in the biomedical domain, and we adopt the \emph{biobert-base-cased-v1.1} as the encoder. The maximum span length $L$ is set as 8. The $r$ is set as 0.05. See Appendix \ref{app:exp_setup} for additional experimental settings.
We use the same combination of hyperparameters for all experiments, and the reported results are the average of 5 runs with different random seeds.

\paragraph{Baselines}
\textsl{KB Matching} retrieves the entities based on string matching with knowledge bases. 
\textsl{AutoNER} \cite{Shang2018LearningNE} filters the distantly annotated datasets through additional rules and dictionaries, and they also proposed a new tagging scheme for the DS-NER task. 
\textsl{Bond} \cite{bond} proposed a two-stage approach to adopt self-training  to alleviate the noisy and incomplete distantly annotated training datasets.
\textsl{bnPU} \cite{peng-etal-2019-distantly} formulates the task as a positive unlabelled learning problem with having the mean absolute error as the objective function.
\textsl{Conf-MPU}~\cite{zhou-etal-2022-distantly} is a two-stage approach, with the first stage estimating the confidence score of being an entity and the second stage incorporating the confidence score into the positive unlabelled learning framework.
\textsl{Span-NS} \cite{li2021empirical} and \textsl{Span-NS-V} \cite{li-etal-2022-rethinking} are the negative sampling approaches, while the latter replaces the previous uniform sampling distribution with a weighted sampling distribution.

As discussed by \citet{zhou-etal-2022-distantly}, the iterative self-training strategy \cite{bond, zhang:2021, DBLP:conf/emnlp/0001ZHWZJ021} could be considered as a post-processing technique that is orthogonal to the single-stage approach. 
We consider the discussion of the self-training \cite{DBLP:conf/nips/ZophGLCLC020} approach beyond the scope of this paper.


\paragraph{Experimental Results}
Table \ref{tab:main_results} shows the comparisons of our approach with the baseline methods on four datasets. 
Our model consistently outperforms the previous approaches in terms of the $F1$ score.
\emph{AutoNER} achieves good performance on the BC5CDR dataset by mining the phrases with external in-domain knowledge, but it does not show similar performance on the other three datasets.
When comparing to the strong baseline \emph{Conf-MPU}, our Top-Neg \textsubscript{BERT} achieves performance improvement of 0.92 and 3.17 F1 points on CoNLL03 and BC5CDR respectively. 
Note that the \emph{Conf-MPU} also reported the results with lexicon feature engineering in the original paper, but they are not directly comparable with our approach.
Our Top-Neg \textsubscript{BERT} also outperforms the previous sampling approach \emph{Span-NS-V} by 1.52 $F1$ points on average.
As the distantly supervised datasets are noisy in terms of both positive and negative samples, the \emph{Span-NS-V} may not have a good classifier to determine the sampling probabilities.
By contrast, our method only relies on the encoded representations of the samples to derive the similarity score for sampling. 
We also conduct experiments with RoBERTa as the encoder so as to have a fair comparison with the \emph{BOND}.
When the stronger pre-trained model is applied to our approach, we observe better performance on all datasets.

\begin{table*}[!t]
    \centering
    \resizebox{1\textwidth}{!}{
    \begin{tabular}{l@{\hspace{0.23cm}}lcccccccccccc}
    \toprule
    \multirow{2}{*}{\textbf{Mode}}  & \multirow{2}{*}{\textbf{Model}}  & \multicolumn{3}{c}{ \textbf{CoNLL03}} & \multicolumn{3}{c}{ \textbf{BC5CDR}}  & \multicolumn{3}{c}{ \textbf{WNUT16}}  & \multicolumn{3}{c}{ \textbf{WikiGold}} \\ \cmidrule(lr){3-5} \cmidrule(lr){6-8} \cmidrule(lr){9-11}  \cmidrule(lr){12-14}
     & & $P.$ & $R.$ & $F1$& $P.$ & $R.$ & $F1$& $P.$ & $R.$ & $F1$ & $P.$ & $R.$ & $F1$ \\
    \midrule
    \textbf{FS} & {Existing SOTA} & - & - & 94.60$^\clubsuit$ & - & - & 90.99$^\blacklozenge$ & - & - & 58.98$^\blacklozenge$ & 62.25 & 66.12 & 64.13$^\spadesuit$ \\
    \midrule
    \multirow{9}{*}{\textbf{DS}} &  KB Matching\textsuperscript{*} & 63.75 & 81.13 & 71.40 &  51.24 & 86.39 & 64.32$^{\dag}$  & 32.22 & 40.34 & 35.83 & 47.63 & 47.90 & 47.76  \\
    &AutoNER \cite{Shang2018LearningNE}\textsuperscript{*} & 60.40 & 75.21 & 67.00  & 77.52 & 82.63 & 79.99$^{\dag}$  & 18.69 & 43.26 & 26.10 & 52.35 & 43.54 & 47.54  \\
    &BOND\textsubscript{RoBERTa} \cite{bond}&  68.90 & 83.76 & 75.71  & - & - & - & 41.52 & 53.11 & 46.61 & 54.40 & 49.17 & 51.55  \\ 
    &bnPU \cite{peng-etal-2019-distantly}$^{\dag}$  &  82.97 & 74.38 & 78.44 &  77.06 & 48.12 & 59.24 & - & - & -  & - & - & -  \\
    &Conf-MPU  \cite{zhou-etal-2022-distantly}$^{\dag}$  &  79.75 & 78.58 & 79.16  & 86.42 & 69.79 & 77.22 & - & - & -  & - & - & -  \\
    &Span-NS \cite{li2021empirical}$^\ddag$ & 80.41 & 71.35 & 75.61 & 86.90 &  73.49 & 79.64  & 53.51 & 39.76 & 45.62 &  51.05 & 48.27 & 49.62\\
    &Span-NS-V \cite{li-etal-2022-rethinking}$^\ddag$ & 80.19 & 72.91 & 76.38 &  86.67 & 73.52 & 79.56 & 47.78 & 44.37 & 46.01 & 50.91 & 48.43 & 49.64\\
    \cmidrule(lr){2-14}
    &\textbf{Top-Neg} (BERT) & 82.72 & 77.71 & 80.08  & - & - & - & 55.28 & 40.35 & 46.55 & 55.47 & 48.57 & 50.65 \\
    &\textbf{Top-Neg} (RoBERTa) & 81.07 & 80.23 & \textbf{80.55}  & - & - & -  & 60.55 & 45.33 & \textbf{51.78} & 52.30 & 53.55 & \textbf{52.86 }\\
    &\textbf{Top-Neg} (Bio-BERT) &- & - & -  & 82.09 & 78.90 & \textbf{80.39 }&- & - & - &- & - & -  \\
    \bottomrule
    \end{tabular}
    }
    \caption{Experiment results. `FS'' and ``DS'' indicate fully supervised and distantly supervised respectively. The existing SOTA results marked with $^\clubsuit$  are retrieved from~\cite{wang-etal-2021-automated}, $^\blacklozenge$ are from  \cite{wang2021improving} and $^\spadesuit$ are from \cite{zhang:2021}. 
    The results with $^*$ are retrieved from \cite{bond}, and the results with $^{\dag}$ are retrieved from \cite{zhou-etal-2022-distantly}. $\ddag$ indicates the results of our runs with their released code.  See Appendix \ref{app:addtional_results} for the standard deviation of our results based on 5 different runs and also the results on the development sets. } 
    \label{tab:main_results}
\end{table*}

\section{Analysis}
\paragraph{Comparison with human-annotated training data}
\label{ana:gap}
We compare the performance of our \emph{Top-Neg} with the standard span-based model\footnote{This span-based model uses all the negative samples.} on the human-annotated (HA) and distantly supervised~(DS) training sets in Table \ref{ana_table:training}.
When the HA dataset is used, our \emph{Top-Neg} achieves comparable performance with the standard \emph{Span} approach.
This demonstrates that using all the negative samples for training is unnecessary.
However, when the noisy DS dataset is used, the performance of the \emph{Span} approach degrades significantly, especially the recall score. 
Our \emph{Top-Neg} approach achieves better performance with relatively balanced precision and recall scores by sampling the effective negatives.
Additionally, the performance gap of our approach on the HA and DS datasets indicates the room to further differentiate the true negatives from false negative samples.

\begin{table}[t]
\centering
\resizebox{0.85\columnwidth}{!}{
\begin{tabular}{lcccc}
\toprule
    \textbf{Model} & Training & $P.$ & $R.$ & $F1$  \\
\midrule
     Span &    HA  & 91.14 & 91.68 & 91.41 \\
     \textbf{Top-Neg} & HA  & 91.48 & 91.66 & 91.57 \\
    \midrule
    Span & DS & 88.25 & 63.03 & 73.54 \\
    \textbf{Top-Neg} & DS  & 82.72 & 77.71 & 80.08  \\
\bottomrule
\end{tabular}
}
\caption{Comparisons on human-annotated (HA) and distantly supervised (DS) training data of CoNLL03.}
\label{ana_table:training}
\end{table}

\paragraph{Comparison of sampling strategies}
As mentioned, we propose to differentiate the false negatives from the true negatives based on the similarity between the negative sample with all positive samples. 
We conduct additional evaluations to show the effect of different sampling strategies on the performance, and Table \ref{table:rates} shows the comparisons.
First, we compare the performance of our approach when only selecting the negative samples with the top similarity score (Eq. \ref{eq:sim}). 
We observe that the performance of our \emph{Top-Neg} with 3\% of the negative samples is worse than 5\%. This indicates that the top 3\% of negative samples are not adequate for training. However, when more negatives are selected (10 \%), we observe a significant drop in the recall score as the number of false negative samples could become dominant. 
By contrast, when the negative samples with low similarity scores~$\Phi$ are selected, the performance shows a significant decrease (lower section of Table \ref{table:rates}). 
Even though the low similarity score indicates a high probability of being a true negative sample, however, these negative samples are less informative.\footnote{See Appendix \ref{app:addtional_results_ha} for the experiment results of using different sampling strategies on the HA dataset. }

\begin{table}[t]
\centering
\resizebox{.8\columnwidth}{!}{
\begin{tabular}{lccc}
\toprule
    \textbf{Sampling}  & $P.$ & $R.$ & $F1$  \\
\midrule
    Top 3\% & 84.73 & 78.61 & 81.55\\
    Top 5\% & 85.78 & 79.38 &	82.42  \\
    Top 10\% & 88.12 & 75.51 & 81.33\\
    \midrule
    Bottom 90\% &  75.33 & 70.82 & 73.01 \\
    Bottom 95\% & 80.36 & 69.00 & 74.25 \\
    Bottom 97\% & 82.92 & 71.96 & 77.05 \\
\bottomrule
\end{tabular}
}
\caption{Results on the development set of CoNLL03 with different sampling strategies. }
\label{table:rates}
\end{table}

\section{Conclusion}

In this work, we propose an improved approach of sampling the negatives to reduce the number of false negative samples for training. Specifically, we differentiate the true negatives from the false negative samples by measuring the similarity of the negatives with the positive samples.
The experiment results have demonstrated the effectiveness of our approach.
Future work may focus on clustering the negative samples to further differentiate the true negatives from the false negatives.

\section*{Limitations}
Our approach is proposed based on the intuition that false negative samples  should have high similarities with the positive samples that have the same gold entity type, and they also have low similarities with the positive samples of different entity types.
However, our proposed approach does not guarantee the selected negatives are true negatives.
Furthermore, when the negative samples are hard false negative samples, they are likely to have high similarities with other positive samples as well.
However, such hard false negative samples are not prevalent in the datasets.
Another limitation is that there is still a large performance gap between the distantly supervised datasets and the human-annotated datasets, as mentioned in Section~\ref{ana:gap}. 



\bibliography{acl2023}
\bibliographystyle{acl_natbib}

\appendix

\section{Dataset Statistics}
\label{app:stats}
Table \ref{app_table:stats} shows the evaluation results of the distantly supervised annotation when compared with human-annotated datasets. We observe that the results often show high precision but low recall scores. 

Table \ref{tab:stats} presents the statistics of the four distantly annotated datasets. ``\# Sent.'' indicates the number of sentences and ``\# Entity'' denotes the number of entities in the datasets.
The training set is annotated based on the distant supervision, and the development and test sets are manually annotated.

\begin{table}[h]
\centering
\resizebox{0.8\columnwidth}{!}{
\begin{tabular}{llccc}
\toprule
    \textbf{Datasets} & Type & $P.$ & $R.$ & $F1$  \\
\midrule
    \multirow{4}{*}{ \textbf{CoNLL03} } & PER &   \textcolor{white}{0}82.36 & 82.11 & 82.23 \\
    & LOC &  \textcolor{white}{0}99.98 & 65.20 & 78.93 \\
    & ORG & \textcolor{white}{0}90.47 & 60.59 & 72.57 \\
    & MISC & 100.00 & 20.07 & 33.43 \\
\midrule
    \multirow{2}{*}{ \textbf{BC5CDR} } & Chemical &  \textcolor{white}{0}96.99 & 63.14 & 76.49 \\
    & Disease &  \textcolor{white}{0}98.34 & 46.73 & 63.35 \\
\bottomrule
\end{tabular}
}
\caption{Evaluation results of the distantly annotated datasets based on human annotation.}
\label{app_table:stats}
\end{table}

\section{Preliminary Experiment Result}
\label{app:preliminary}
\begin{figure}[h]
\centering
\begin{tikzpicture}[scale=0.64]
\pgfplotsset{width=6.5cm, height=5cm, compat=1.3}
\begin{axis}[
    title style={anchor=north,yshift=-0.3cm},
    title = Precision,
    ymin=30, ymax=118,
    xmin=0, xmax=16,
    xlabel={Epoch},
	legend pos = south east,
    ]
    \addplot [mark=o, mark size=1pt, color=skyblue] plot coordinates {
    (1, 83.37)
    (2, 87.711)
    (3, 93.65)
    (4, 94.85900000000001)
    (5, 97.271)
    (6, 98.68400000000001)
    (7, 99.44800000000001)
    (8, 99.132)
    (9, 99.583)
    (10, 99.72999999999999)
    (11, 99.579)
    (12, 99.803)
    (13, 99.77000000000001)
    (14, 99.63)
    (15, 99.837)
    };
    \addplot [mark=triangle, mark size=1pt, color=orange] plot coordinates {
    (1, 92.67999999999999)
    (2, 89.792)
    (3, 86.272)
    (4, 83.21499999999999)
    (5, 83.687)
    (6, 80.694)
    (7, 84.17999999999999)
    (8, 82.947)
    (9, 85.721)
    (10, 84.22699999999999)
    (11, 82.56400000000001)
    (12, 84.112)
    (13, 83.654)
    (14, 82.982)
    (15, 84.49600000000001)
    };
\legend{Train \\ Dev\\}
\end{axis}
\end{tikzpicture}
\begin{tikzpicture}[scale=0.64]
\pgfplotsset{width=6.5cm, height=5cm, compat=1.3}
\begin{axis}[
    title style={anchor=north, yshift=-0.3cm},
    title = Recall,
    ymin=30, ymax=118,
    xmin=0, xmax=16,
    xlabel={Epoch},
	legend pos = south east,
    ]
    \addplot [mark=o, mark size=1pt, color=skyblue] plot coordinates {
    (1, 73.92699999999999)
    (2, 86.345)
    (3, 93.386)
    (4, 97.762)
    (5, 98.628)
    (6, 99.095)
    (7, 98.22800000000001)
    (8, 99.533)
    (9, 99.438)
    (10, 99.775)
    (11, 99.786)
    (12, 99.736)
    (13, 99.843)
    (14, 99.949)
    (15, 99.933)
    };
    \addplot [mark=triangle, mark size=1pt, color=orange] plot coordinates {
    (1, 61.797000000000004)
    (2, 64.69200000000001)
    (3, 62.504000000000005)
    (4, 65.33200000000001)
    (5, 63.715999999999994)
    (6, 60.702999999999996)
    (7, 61.074)
    (8, 63.278)
    (9, 61.931999999999995)
    (10, 62.370000000000005)
    (11, 62.639)
    (12, 61.831)
    (13, 61.41)
    (14, 64.339)
    (15, 62.370000000000005)
    };
\legend{Train \\ Dev\\}
\end{axis}
\end{tikzpicture}
\caption{Precision (\%) and recall (\%) on the training and development sets of CoNLL03. Note that the best performance (F1 score) on the development set is at the 2nd epoch. }
\label{fig:ana1}
\end{figure}
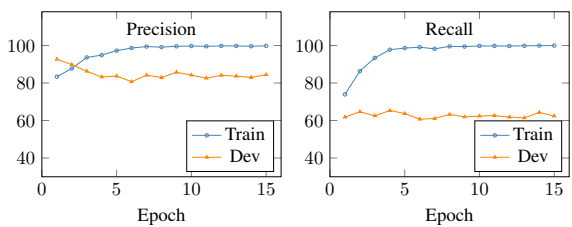

Figure \ref{fig:ana1} shows the experiment results of a standard span-based model that is trained on the distantly annotated CoNLL03 dataset. The evaluation is conducted on the human-annotated development and test sets.

\section{Additional Experimental Setup}
\label{app:exp_setup}
We use the \emph{bert-base-cased} and \emph{roberta-base} as the encoders for CoNLL03, WNUT16, and WikiGold datasets. BC5CDR is in the biomedical domain, and we adopt the \emph{biobert-base-cased-v1.1} as the encoder. We use 2 layers of feed-forward neural networks for the classifier and the hidden size is set as 150, and the dropout rate is set as 0.2. The maximum span length $L$ is set as 8. The $r$ is set as 0.05. We use the same combination of hyperparameters for all the experiments. We select the best model based on the performance on the development sets, and the reported results are the average of 5 runs with different seeds.

The experiments are conducted on Nvidia Tesla A100 GPU with PyTorch 1.10.0. The average running time on the CoNLL03 dataset is 74 seconds/epoch, and the number of model parameters is 108.59M when \emph{bert-base-cased} is adopted.

\begin{table}[t]
\centering
\resizebox{\columnwidth}{!}{
\begin{tabular}{l@{\hspace{0.13cm}}lcccc}
\toprule
    \multicolumn{2}{l}{\textbf{Datasets}} & CoNLL03  & BC5CDR & WNUT16 & WikiGold \\
\midrule
    \multirow{2}{*}{ \textbf{Train} } & \# Sent. & 14,041  & 4,560 & 2,393 & 1,142\\
    & \# Entity & 17,781  & 6,452 & \textcolor{white}{0,}994 & 2,282\\
\midrule
    \multirow{2}{*}{ \textbf{Dev} } & \# Sent. & \textcolor{white}{0}3,250  & 4,579 & 1,000 & \textcolor{white}{0,}280 \\
    & \# Entity & \textcolor{white}{0}5,942  & 9,591 & \textcolor{white}{0,}661 & \textcolor{white}{0,}648\\
\midrule
    \multirow{2}{*}{ \textbf{Test} } & \# Sent. & \textcolor{white}{0}3,453  & 4,797 & 3,849 & \textcolor{white}{0,}274\\
    & \# Entity & \textcolor{white}{0}5,648  & 9,809 & 3,473 & \textcolor{white}{0,}607\\
\bottomrule
\end{tabular}
}
\caption{Statistics of datasets. }
\label{tab:stats}
\end{table}

\begin{table}[t]
\centering
\resizebox{\columnwidth}{!}{
\begin{tabular}{l@{\hspace{0.13cm}}cccc}
\toprule
    \textbf{Datasets} & CoNLL03  & BC5CDR  & WNUT16 & WikiGold \\
\midrule
    \textbf{Top-Neg} \textsubscript{BERT} & 82.42  &  - & 44.34 & 55.10  \\
    \textbf{Top-Neg} \textsubscript{RoBERTa} &  83.67 & -  &  48.78 &  57.46     \\
    \textbf{Top-Neg} \textsubscript{BioBERT} & -  & 80.69 & - & - \\
\bottomrule
\end{tabular}
}
\caption{Experiment results on the development sets. }
\label{app_table:dev_results}
\end{table}

\begin{table}[h]
\centering
\resizebox{\columnwidth}{!}{
\begin{tabular}{l@{\hspace{0.13cm}}cccc}
\toprule
    \textbf{Datasets} & CoNLL03  & BC5CDR  & WNUT16 & WikiGold \\
\midrule
    \textbf{Top-Neg} \textsubscript{BERT} & 0.94  & - & 1.09 & 1.03 \\
    \textbf{Top-Neg} \textsubscript{RoBERTa} & 0.63 & - & 0.42 & 0.25  \\
    \textbf{Top-Neg} \textsubscript{BioBERT} & -  & 0.32 & - & - \\
\bottomrule
\end{tabular}
}
\caption{Standard deviation of the $F1$ score on the test sets. }
\label{app_table:variance}
\end{table}

\section{Additional Experiment Results}
\label{app:addtional_results}
In this section, we show additional experiment results. Table \ref{app_table:dev_results} presents the results of our approach on the development sets of the four datasets. As mentioned that we run our model with different seeds for 5 times, Table \ref{app_table:variance} shows the standard deviation of the $F1$ scores on the test sets.

\section{Additional Experiment Results on the HA Dataset}
\label{app:addtional_results_ha}
Table \ref{table:rate_ha} shows the experimental results on the development set of CoNLL03 when using different sampling methods on the HA dataset.
The experiment with the top 5\% of the negative samples achieves comparable performance when using all the negative samples.
We observe a large performance gap between the settings of the top 5\% and bottom 5\%. This indicates that the bottom negative samples are less informative than the top negatives.
When the bottom 50\% of negative samples are selected, the performance shows improvement, but it still exists a gap when compared with the top 50\%.

\begin{table}[h]
\centering
\resizebox{.6\columnwidth}{!}{
\begin{tabular}{lccc}
\toprule
    \textbf{Sampling}  & $P.$ & $R.$ & $F1$  \\
\midrule
    ALL & 95.65 & 95.93 & 95.79 \\
    \midrule
    Top 5\% &  95.70 & 95.76 & 95.73 \\
    Bottom 5\% & 64.85 & 96.90 & 77.70 \\
    \midrule
    Top 50\% &  96.10 & 95.31 & 95.70 \\
    Bottom 50\% & 89.70 & 96.13 & 92.80  \\
\bottomrule
\end{tabular}
}
\caption{Comparisons of different sampling strategies on the development set of the HA CoNLL03. }
\label{table:rate_ha}
\end{table}
\end{document}